\documentclass{article}

\usepackage[export]{adjustbox}
\usepackage{amsmath}
\usepackage{amssymb}
\usepackage{amsthm}
\usepackage{bm}
\usepackage{booktabs}
\usepackage{caption}
\usepackage{enumitem}
\usepackage{float}
\usepackage[T1]{fontenc}
\usepackage{graphicx}
\usepackage{hyperref}
\usepackage[utf8]{inputenc}
\usepackage{makecell}
\usepackage{mathtools}
\usepackage{microtype}
\usepackage{multirow}
\usepackage{pifont}
\usepackage{subcaption}
\usepackage{soul}
\usepackage[most]{tcolorbox}
\usepackage{url}
\usepackage{wrapfig}
\usepackage{pgf-pie}
\usepackage{icml2026/algorithm}
\usepackage{icml2026/algorithmic}

\newcommand\freefootnote[1]{%
  \let\thefootnote\relax%
  \footnotetext{#1}%
  \let\thefootnote\svthefootnote%
}

\usepackage[accepted]{icml2026/icml2026}

\usepackage[capitalize,noabbrev]{cleveref}

\icmltitlerunning{Want Better ML Reviews? Stop Asking Nicely and Start Incentivizing}

\begin{document}

\twocolumn[
  \icmltitle{Position: Want Better ML Reviews?\\ Stop Asking Nicely and Start Incentivizing with a Credit System}

  \icmlsetsymbol{equal}{*}
  \begin{icmlauthorlist}
    \icmlauthor{Shaochen (Henry) Zhong}{rice}
  \end{icmlauthorlist}
  \icmlaffiliation{rice}{Department of Computer Science, Rice University, Houston, TX, USA}
  \icmlcorrespondingauthor{Shaochen (Henry) Zhong}{henry.zhong@rice.edu}

  \icmlkeywords{Machine Learning, ICML}

  \vskip 0.3in
]

\printAffiliationsAndNotice{}  


\begin{abstract}

With soaring submission counts, stricter reciprocal review policies, widespread adoption of platforms like OpenReview, and without the offsetting pressure of publication fees, the machine learning (ML) community has one of the largest scholarly presences among all scientific fields. And yet, \textbf{almost \textit{everyone} has \textit{many} unpleasant things to share about their review experience.} Worse, there is little public space to seriously discuss, let alone debate, what makes a review system effective or how it might be improved.\quad In this position paper, we expand our discussion from two core problems: \textit{How can we reasonably limit submission volume?} and \textit{How can we incentivize good and discourage bad reviewing?} We first assess the strengths and shortcomings of existing attempts to address such problems. Specifically, we present four takes on some popular conference mechanisms and propose two alternative designs for improvement.\quad Our general position is that meaningful improvement in ML peer review won't come from polite best-practice suggestions tucked into Calls for Papers or Reviewer Guidelines: it requires \textbf{enforceable yet fine-grained procedural safeguards} paired with \textbf{a currency-like credit system (e.g., our proposed \textit{OpenReview Points})}. ML practitioners can ``earn'' such points by contributing good review practices, and ``spend'' them across one or multiple major conferences to redeem different kinds of ``perks,'' such as complimentary registration or the right to request additional review resources.

\end{abstract}

\section{Introduction}

\textbf{This position paper argues that peer review in machine learning (ML) is unlikely to improve through polite requests or optimistic guidance tucked into Calls for Papers or Reviewer Guidelines. Fine-grained yet enforceable procedural guardrails, combined with a spendable, across-conference credit system, are almost mandatory for a sustainable review ecosystem.}

Machine learning has scaled faster than nearly any other scientific field in both volume and visibility. We now have tens of thousands of paper submissions to a single conference,\footnote{The most recent \href{https://blog.neurips.cc/2025/09/30/reflections-on-the-2025-review-process-from-the-program-committee-chairs/}{NeurIPS 2025} had 21,575 valid submissions to the main conference alone.} open-access platforms like OpenReview that support interactive discussions, and increasingly reciprocal reviewing obligations to match supply with demand. On paper, the ML community has everything it needs to sustain a robust yet pleasant peer-review pipeline: we have the largest scholarly presence and the most modern review technology, all without the typical bottlenecks of paywalls, publication fees, or expensive memberships. However, the lived reality often feels far less functional. From cryptic or dismissive reviews to wildly inconsistent standards, frustrations with the review process appear widely shared \citep{Jakobsen2022peer_review_factors}, voiced by PhD students, seasoned professors, and industry researchers alike (e.g., see examples in Section~\ref{sec_root_num}).\footnote{In the \href{https://blog.neurips.cc/2025/12/05/neurips-datasets-benchmarks-track-from-art-to-science-in-ai-evaluations/}{NeurIPS 2025 Datasets \& Benchmarks author survey}, for instance, roughly 25\% of respondents flagged review quality as needing improvement.} Worse, there is little to no structured way to hold bad actors accountable, nor are there incentives to encourage good actors to go the extra mile.



In this position paper, we expand our discussion of the two core challenges we have identified:

\begin{enumerate}[leftmargin=*, noitemsep, topsep=0pt]
    \item \textbf{\textit{How can we reasonably limit submission volume?}}
    \item \textbf{\textit{How can we incentivize good and discourage bad reviewing?}}
\end{enumerate}

We first lay the background on why these two issues are the root causes of much unpleasantness in ML review. Then, we assess some existing attempts to mitigate such issues as implemented in several ML conferences. We present our takes on such measures and, finally, propose two new mechanisms: \textbf{fine-grained procedural safeguards that could be enforced at scale; and a credit system based on something we call \textit{``OpenReview Points''}, which would let researchers ``earn'' and ``spend'' their reviewing efforts in tangible ways across major conferences and review cycles.} We believe such mechanisms would have a fair chance of addressing many of the aforementioned shortcomings effectively and, more importantly, are flexible enough to allow each conference to adopt its own variants. Beyond our proposed mechanism, we engage many alternative views, where we discuss how such views are valid (or not) and how our proposed mechanism shall be able to take such concerns into consideration. We conclude our paper with a \textit{Recommended Practices} section, which outlines our vision on how the first few conferences adopting a similar credit system should proceed and what aspects should be considered cautiously. Additional materials, such as anecdotal case studies based upon real conferences, can be found in Appendix~\ref{app_case_study}. Due to limited immediately relevant work and page limits, we present related works in Appendix~\ref{app_related}.

We emphasize that our goal is not to perfect ML peer review (as it would be unfaithful and condescending for anyone to claim so), but to make its failures rarer, less painful, and, most importantly, more accountable and sustainable. We are also not here to propose a specific rulebook that all conferences must follow, but rather to advocate for a promising general direction that future conference organizers can explore and adapt to their own needs.



\section{Root Causes}
\label{sec_root}

\subsection{Overwhelming number of submissions causes all kinds of challenges.}
\label{sec_root_num}

We believe it is common knowledge that ML conferences typically receive an overwhelming number of submissions \citep{kim2025reviewer_rewards, yang2025paper_copilot}, driven by the field's rapid growth and increasingly accessible AI-assisted research (Section~\ref{sec_takes_caps}). Naturally, this causes all kinds of practical challenges. From a manpower perspective, more submissions directly mean greater demand for reviewers and Area Chairs (ACs), which translates to a heavier workload for Senior Area Chairs (SACs) and, eventually, Program Chairs (PCs). With such great pressure on every aspect of the conference review system, the results are almost predictable: thinner attention per paper, more rushed triage, and greater variance in both review quality and decision outcomes \cite{su2025self_rank, shah2022challenges_pr}.

Moreover, practicality-wise, most ML conferences often guarantee the right to in-person presentation exposure once a paper is accepted.\footnote{That said, this tradition might be undergoing a serious update, with \href{https://icml.cc/Conferences/2026/CallForPapers}{ICML 2026} no longer requiring in-person attendance.} This makes physical capacity restrictions come into play, directly imposing an upper limit on how many papers can be accepted. Many borderline or acceptance-inclined papers might be ruled out purely based on capacity constraints, a role often delegated to SACs. But considering the number of submissions versus the number of SACs,\footnote{In \href{https://media.neurips.cc/Conferences/NeurIPS2024/NeurIPS2024-Fact_Sheet.pdf}{NeurIPS 2024}, there were 195 main-conference SACs against 15,671 submissions, roughly 80 papers per SAC.} this kind of assignment is, by design, unreasonable and unsustainable, as few SACs would have the bandwidth or appetite to go through the content and review record of that many papers. In practice, this pressure incentivizes shortcutting (e.g., relying more heavily on numerical scores or other similar quick heuristics), which further amplifies randomness and weakens accountability. In fact, we have seen many SACs publicly pushing against such ``force rejection for capacity'' practices, as exemplified by LinkedIn posts from NeurIPS SACs \href{https://www.linkedin.com/posts/abeirami_my-thoughts-on-the-broken-state-of-ai-conference-activity-7375191053093728256-Jxdv}{Ahmad Beirami} and \href{https://www.linkedin.com/posts/atlas-wang-41b726259_neurips-activity-7374496205533466624-iDjl}{Atlas Wang}.

\subsection{Lack of oversight, feedback loops, and incentives for good actors and consequences for bad ones.}
\label{sec_root_feedback}

While reviewers, ACs, and SACs have the right to provide feedback, ML conferences lack proper oversight and feedback loops. A reviewer can act with near-total impunity — submitting dismissive, inconsistent, or low-effort assessments — so long as they are not extreme enough to trigger formal intervention, and little exists to correct or even surface such behavior. \textbf{This ecosystem leaves actors with few avenues to learn how to become better, let alone much incentive to go the extra mile.} Without routinized feedback, transparent metrics, or positive incentives, the system neither rewards exemplary stewardship nor deters poor practices; we elaborate on this in Section~\ref{sec_implicit_expectations}.

\section{Why Existing Fixes Fall Short}

\subsection{Soft and hard submission caps offer limited help.}
\label{sec_takes_caps}

With the growing body of research in the ML community \citep{yang2025paper_copilot, kim2025reviewer_rewards} and with AI-assisted research becoming more accessible\footnote{This is particularly evidenced by recent automatic research tooling such as Karpathy's \href{https://github.com/karpathy/autoresearch}{\texttt{autoresearch}}, and by coding-assistant plans such as \href{https://x.com/bcherny/status/2032514807388123255}{Claude Code} moving Opus's 1M-token context from extra usage into the standard subscription, both lowering the access barrier to AI-assisted work. We can even find entirely AI-driven work accepted at major NLP venues, e.g., an AI scientist getting a paper into the main track of ACL 2025 (see \href{https://www.lesswrong.com/posts/LtsgfGsXpiLTSGpaW/zochi-publishes-a-paper?utm_campaign=post_share&utm_source=link}{this blog} and \citealp{zhou2025tempest}).} \citep{eger2025ai_science_survey}, the volume of submissions continues to grow at a pace that far outstrips the community's reviewing capacity. This escalation naturally prompts discussions around mechanisms for curbing submission rates and maintaining a manageable reviewing load, where submission caps are often proposed as one of the most direct ways to reduce such volume.

\begin{table}[H]
\centering
\small
\caption{\textbf{Most-submitted authors at ICLR 2025, counted by author position} per PaperCopilot statistics \citep{yang2025paper_copilot}. Each cell reports the submission count of the $n$-th most prolific author under that counting. Notably, the top \emph{any author} counts far exceed the top \emph{first} or \emph{last author} counts.}
\label{tab_papercopilot}
\begin{tabular*}{\columnwidth}{@{\extracolsep{\fill}}l|ccccc@{}}
\toprule
Most submitted & \#1 & \#25 & \#50 & \#75 & \#100 \\
\midrule
First author & 7  & 4  & 3  & 3  & 3  \\
Last author  & 34 & 13 & 11 & 9  & 8  \\
Any author   & 42 & 21 & 17 & 16 & 14 \\
\bottomrule
\end{tabular*}
\end{table}
\vspace{-1em}

We argue that submission caps, whether ``soft'' (e.g., mandatory reciprocal review beyond a certain number of submissions) or ``hard'' (e.g., strict per-author quotas on how many papers can be submitted), provide, at best, marginal relief. The core problem is not that a small set of ``hyper-prolific'' lead authors are personally flooding the system with many new submissions \citep{yang2025paper_copilot}; rather, it is distributed across the community, as there is essentially no immediate downside for any author to submit unready manuscripts or endlessly recycle previously rejected work with critical flaws, straining reviewing capacity in aggregate.

We suspect that, in practice, per-author caps mostly trim auxiliary authors from the byline so that teams can fit under the quota. In other words — without penalties for low-quality submissions — \textbf{submission caps may largely change who gets listed on a paper, rather than whether the paper is submitted.} While we lack direct evidence, public statistics are at least suggestive: as shown in Table~\ref{tab_papercopilot}, a non-trivial number of authors are listed on many ICLR 2025 submissions, yet very few are \textit{first} or \textit{last} authors on comparably many, making it unlikely that they are core contributors to all those submissions. Under a harsh cap, such authors would likely remove their names from lower-priority submissions rather than withhold those papers from submission altogether. Thus, we argue that until there is a genuine negative incentive that discourages unlimited resubmission and/or rewards restraint, submission caps can only nibble at the edges of the volume problem, instead of making a significant impact at scale.

\subsection{Irresponsible Reviewers Care Most About Their Own Works, So Asking Nicely is Not Helpful}
\label{sec_takes_reviewers}

Bad or irresponsible review practices appear widespread, in large part because of the lack of accountability built into current conference mechanisms. Until very recently, most ML conferences enforced no sanction against irresponsible reviewers, leaving bad practices essentially unchecked. For a reviewer who is recruited by force (e.g., through mandatory reciprocal reviewing) and treats the assigned duty as a mere task to be finished, the things that matter most are likely their own current or future submissions. We argue that, \textbf{to effectively discourage irresponsible reviewing, some form of sanction must be enforced at the submission end.} Otherwise, conferences have little real leverage and often can only resort to asking nicely in Calls for Papers or Reviewer Guidelines; and despite the existence of some extremely thoughtful guidelines like the \href{https://aclrollingreview.org/reviewerguidelines}{ARR Reviewer Guidelines}, their effectiveness leaves much to be desired.

Recently, starting with CVPR 2025, several ML conferences have adopted what is essentially a retaliatory desk-rejection policy targeting irresponsible reviewers. At CVPR 2025, its Area Chairs (ACs) \textit{``identified a number of highly irresponsible reviewers, those who either abandoned the review process entirely or submitted egregiously low-quality reviews, including some generated by large language models''} and ultimately issued \href{https://x.com/CVPR/status/1894853624200863958}{desk rejections for 19 otherwise accepted papers} involving those reviewers.

While this act marks a meaningful start to enforcing hard procedural guardrails to protect review quality and integrity,\footnote{E.g., at \href{https://blog.icml.cc/2026/03/18/on-violations-of-llm-review-policies/}{ICML 2026}, watermark detection flagged roughly 1\% of all reviews as LLM-generated despite a no-LLM policy, and about 2\% of submissions were desk-rejected over such violations — both are small numbers. We also expect such a mechanism to grow less fruitful as awareness of these watermarks spreads.} we argue that \textbf{retaliatory procedures as harsh as desk rejection can only offer marginal benefits to the conference at large, as only a few bad actors would be extreme enough to blatantly ignore direct instructions while leaving indisputable evidence behind.} This is because a harsh penalty like desk rejection is, by nature, a blunt instrument: it is too harsh to apply broadly and can only be reasonably used for the most extreme violations with verifiable signals. In CVPR's case, it was mostly reserved for reviewers who outright abandoned their review duties, a clean, rule-based, verifiable breach that leaves no room for ambiguity.\quad Unfortunately, most reviewer problems in ML are arguably more subtle than complete negligence. Irresponsible reviews can manifest in many forms: from thoughtless boilerplate complaints like ``no theory'' or ``needs more experiments'' applied indiscriminately to every submission, to a gross misunderstanding of basic facts and refusal to reconsider, to a raised concern with no concrete support, or even the famous \href{https://x.com/2prime_PKU/status/1948549824594485696}{``Who is Adam?''}. These reviews are much harder to police, but no less damaging.

We argue that desk rejection is too coarse a penalty to handle the long tail of poor reviewing behaviors that fall short of full abandonment. If we want meaningful deterrents at scale, we need a system that applies graduated, proportional penalties, not just all-or-nothing rulings. The fact that the CVPR 2025 procedure only resulted in 19 desk rejections suggests that the irresponsible review issue in the ML community is far from resolved by adopting this retaliatory desk rejection policy alone; more enforceable yet fine-grained procedural safeguards are necessary to handle the wide spectrum of irresponsible review practices. In other words, \textbf{desk rejections are like felony charges, but we also need misdemeanors, infractions, and everything in between to moderate a proper review community.}

\subsection{100\% Mandatory Reciprocal Reviewer Recruitment is a Slow-Acting Poison}

To keep up with rising submission counts, many ML conferences, such as the most recent \href{https://aclrollingreview.org/incentives2025}{EMNLP 2025 (ARR May)}, now rely on 100\% reciprocal reviewer recruitment: every eligible author\footnote{Where such eligibility is determined by having at least a few published works at certain recognized ML/AI venues.} must also be available to review, except in extreme circumstances like parental leave. On paper, this sounds fair: if one wants to publish, one should also contribute to the review pool. But in practice, we argue it is a slow-acting poison that comes at the cost of review quality.

The policy assumes that all eligible authors of every accepted paper are both capable and willing to provide thoughtful reviews. That assumption does not hold across the ML community: many authors may have played only auxiliary roles or contributed as expert consultants on highly specialized components. They are therefore ill-suited to reviewing general-purpose ML submissions.\quad Worse, mandatory review removes the ability for people to decline, even when they know they cannot meaningfully contribute due to sensible (but non-medical) emergencies and circumstances. Once in the reviewer pool, conferences often allow very limited flexibility for exemption. For instance, AAAI 2026 instructed their reviewers to \textit{``do your best''} even if the assigned paper is outside their area of expertise. 

We argue that such enforced cultures would likely result in a series of rushed, templatized, or often disengaged reviews/meta-reviews. We suspect this is a meaningful contributor to much of the unpleasantness in ML peer review: \textbf{when recruited by force and with limited motivation and bandwidth, reviewers are likely to invest only the minimum effort, as their main drive becomes fulfilling their assigned duties so their own submitted work is not desk-rejected.}\quad Perhaps recognizing this, many later conferences appear to be on the same page with us (in terms of realizing the negative side of mandatory reciprocal review): starting from \href{https://aclrollingreview.org/exemptions2025}{ARR July 2025}, technically qualified authors may request an exemption from review duty on a case-by-case basis if they find themselves lacking the relevant expertise. That said, broader allowance for exemption (e.g., lack of bandwidth) has yet to be widely adopted.

We find it ironic that a mechanism designed to distribute the workload ends up degrading its quality. Yet, if everyone could opt out with no consequences, the system would collapse under the sheer volume of submissions. The reasonable middle ground, then, is to allow sensible opt-outs while still holding authors accountable for their share of the reviewing load — ideally alongside a way to mobilize additional willing reviewers at will to cover any resulting gaps. We expand on both in Section~\ref{sec_implicit_expectations} and Section~\ref{sec_proposal}.

\subsection{Helpful Implicit Expectations Often Go Unmet}
\label{sec_implicit_expectations}

As we previously teased in Section~\ref{sec_root_feedback}, conference processes implicitly assume that reviewers, ACs, and SACs will self-initiate best practices, such as timely calibration, substantive internal discussions, careful revision after rebuttals, and principled follow-ups, to ensure that informed decisions are made for each submission. We argue that in reality, \textbf{helpful practices like internal reviewer discussions rarely happen effectively at scale, because reviewers are seldom incentivized to ``go the extra mile.''}

These practices bring little recognition, credit, or accountability for the extra coordination and time they require, so under deadline pressure most actors would likely default to the minimum required. Steps like internal reviewer discussion thus tend to be perfunctory or skipped, and decision quality can degrade, not for lack of guidance, but for lack of aligned incentives to act on it.

As \href{https://www.duperrin.com/english/2014/06/23/quote-tell-how-you-measure-and-i-will-tell-you-how-i-will-behave/}{Eli Goldratt} put it: \textit{``Tell me how you measure me and I will tell you how I will behave''} — when the only thing measured is whether one's assigned duties are formally complete, that is all most will optimize for. We argue \textbf{the key to better review is therefore not to compel more participation, but to find and motivate those with the bandwidth and genuine enthusiasm to engage well — including capable scholars who currently sit out the process entirely.} This is admittedly difficult, as review remains, for now, a largely thankless activity; we explore how to incentivize in Section~\ref{sec_proposal} and Section~\ref{sec_alt_point_exempt}.

\section{Our Proposal: Fine-Grained Procedural Guardrails with a Currency-Like Incentive System}
\label{sec_proposal}

To make meaningful progress in peer review reform, we argue that two ingredients are essential: \textbf{enforceable procedural safeguards at different granularities}, and \textbf{an incentive structure that rewards good-faith participation while offering flexibility.} We propose a system based on a community-wide, cross-conference-supported economy called \textit{``OpenReview Points''}, mainly due to the widespread adoption of OpenReview, which is also well positioned to track such balances.\quad This section outlines the basic principles of such a system, discusses potential enforcement strategies, and explores the feasibility of a conference-wide credit market that could finally provide conference organizers both the ``stick'' and the ``carrot'' they currently lack.

\subsection{OpenReview Points: A Currency-Like Economy Enabling Flexible Options}
\label{sec_proposal_points}

The current review ecosystem operates on the honor system: a reviewer is expected to perform review duties diligently and hope that others will do so as well. However, we argue that optimistic hoping is not a system. To instill accountability, we propose a credit system, giving contributors to the review pipeline something to earn, spend, and track.

Under our proposal, ML practitioners would accumulate OpenReview Points based on their contributions to the community. For instance, in the context of reviewing, completing a standard review might earn 1 point, helping with an emergency review might earn 2, and being recognized as an ``outstanding reviewer'' could grant an additional 3 points.\footnote{We emphasize that all point values mentioned in this section are intuitively assigned for hypothetical purposes. A real OpenReview Point-based economy would require significantly more sophisticated balancing, subject to each conference's own preferences. More on such specifications in Section~\ref{sec_rec}.} Once earned, OpenReview Points could be spent to gain access to certain ``perks'' and privileges. For example:

\begin{itemize}[leftmargin=*, noitemsep, topsep=0pt]
\item A reviewer can spend 5 points to opt out of an assigned review duty.
\item An author can spend 10 points to exempt a co-author from their reciprocal reviewing obligation.
\item An author can spend 50 points to request an additional expert reviewer in the case of a highly controversial or borderline decision. In the meantime, a reviewer/AC can take this job and earn those 50 points.
\item An author can spend 100 points to redeem free registration.
\end{itemize}

This economy introduces direct incentives: if one contributes meaningfully, one gains flexibility and optionality. If one does not, one's publishing privileges will begin to shrink. We emphasize that, \textbf{because credits can be flexibly awarded or deducted for almost any behavior, the system gives organizers a broad design space to influence community behavior in ways never possible} with just blunt-force policies like universal reciprocal reviewing or desk rejections.

On the \emph{reward} side, for instance, much of our work argues that there is a lack of incentive for actors to ``go the extra mile,'' even when such effort can be immensely helpful (e.g., as anecdotally demonstrated in Appendix~\ref{app_case_study}). With point incentives, however, such ``extra miles'' can be explicitly encouraged: reviewers may become more willing to initiate and engage in internal discussion, and ACs more willing to investigate, because exemplary actions can now be potentially rewarded. Similarly, the lack of a reviewer feedback loop, discussed in Section~\ref{sec_root_feedback}, can be mitigated by awarding points to authors who are willing to provide detailed reviewer feedback, which reviewers can consult to improve their future practices.

On the \emph{penalty} side, as noted in Section~\ref{sec_root_num}, many reviewing issues stem from inflated submission counts. \textbf{One way to mitigate this is to require a small and refundable ``submission fee'' (e.g., 10 OpenReview Points) per paper.} If the paper is accepted or meets a reasonable ``fair attempt'' bar, the points are refunded; otherwise, they are forfeited. This soft deterrent discourages unready submissions by linking low-quality or premature work to a corresponding reduction in future publication privileges. We find this use of the credit system particularly appealing, as it directly targets the core issue of submission quantity. It also rests on a relatively reliable signal: the \href{https://blog.neurips.cc/2021/12/08/the-neurips-2021-consistency-experiment/}{NeurIPS Consistency Experiments} found that when two independent reviewer teams assess the same papers, their agreement is strongest on clear rejections. In other words, while review outcomes can be quite noisy for borderline or even spotlight-worthy work, they are far more consistent at flagging low-quality submissions. We envision that such cases can be reliably identified by signals like uniformly low ratings from all reviewers — exactly what a refundable submission fee is meant to discourage.

While we have proposed several specific policies in this section, \textbf{we emphasize that we do not argue for the enforcement of any specific rule}. Rather, we argue that a credit system would grant every participant in the ML community far greater flexibility in how they interact with the review process. While we fully expect friction or disagreement regarding any particular rule or redemption policy, we believe it would be difficult to argue against the utility of having a credit system \emph{at all}, since it makes sense for different conferences to carve out their own rules to cater to their own communities.

\subsection{Making the Credit System Enforceable: A Four-Part Defense}
\label{sec_proposal_defense}

A credit system is only as useful as it is enforceable. Beyond awarding points for good-faith contributions, organizers need levers to deter malicious gaming behaviors, such as fraud, point farming, and bulk low-effort reviewing. Granted the vast scale of potential abuse attempts, these levers should operate at a finer granularity than the blunt instrument of desk rejection. Of course, no system is immune to abuse — even real-world economies with actual laws and tangible consequences face persistent bad actors. Nonetheless, we argue that, although imperfect, a credit system is far better positioned than the status quo, precisely because organizers can award and deduct points to impose countermeasures and realign incentives. We group these countermeasures into four reusable primitives that conferences can mix and match:

\begin{itemize}[leftmargin=*, noitemsep, topsep=0pt]
\item \textbf{Duty Delegation:} Restrict certain actions to specific roles. For example, only a paper's own (lead) authors may spend points to request an additional expert reviewer, and submission-fee-like charges fall on lead authors rather than auxiliary ones. This prevents charges and privileges from being offloaded onto, or routed through, less accountable parties.
\item \textbf{Upper Limits:} Cap how often an action may be taken per cycle, e.g., how many papers one may review for points, or how many additional or emergency reviewer requests one may file. This bounds low-effort point farming and bulk abuse.
\item \textbf{Dynamic Pricing:} Make repeated use of an action or perk progressively more expensive, e.g., beyond a threshold, each submission or each successive additional-reviewer request would cost more points. This reserves scarce resources for the cases that matter most.
\item \textbf{Voting-Based Penalties and Awards:} Let peer reviewers and the AC issue fine-grained, peer-driven judgments, deducting points for low-quality reviews that peers and the AC confirm, and awarding points for exemplary ones (e.g., an ``outstanding reviewer'' bonus).
\end{itemize}

These primitives compose against concrete threats. For instance, bulk outsourcing of reviews for points is curbed by \textit{upper limits} (a few reviews per actor) and \textit{voting-based penalties} (low-quality reviews lose points), making writing good reviews the easier path forward. Schemes that funnel points to a dedicated ``point person'' to be listed as a coauthor across many papers to buy perks are blunted by \textit{duty delegation} (has to be lead authors) and \textit{dynamic pricing} (increasing costs for more perk purchases), and, absent person-to-person transfers, are simply hard to pull off at scale. Similarly, spamming the ``additional expert reviewer'' option is also contained by per-cycle caps and escalating prices. \textbf{While these attack angles represent only a fraction of potential threat models, we argue that the primitives themselves are flexible enough to be adapted to a wide range of abuse attempts}, and that the ability to deploy such countermeasures is a key advantage of a credit system over the status quo.

Crucially, much of this rests on one principle: points must be earned through labor, not bought. We therefore strongly advocate forbidding person-to-person point transfers and any money-to-points conversion. The moment points can be purchased, the system risks degenerating into pay-to-play, which is plausibly worse than the status quo. The voting-based lever in particular raises concerns about false positives and politicization, which we address in Section~\ref{sec_alt_vote}.





\section{Alternative Views}
\label{sec_alt}

While we advocate for enforceable procedural safeguards and a credit system, we recognize that not everyone will agree with this approach. Below, we discuss several alternative perspectives and respond to their concerns.

\subsection{``A credit system gamifies peer review and adds needless bureaucracy.''}

A common objection is that introducing a credit system risks gamifying the review process, turning what should be a scholarly, community-driven responsibility into a transactional system; a related worry is that tracking points and adjudicating review quality would introduce too much bureaucracy. Both concerns are understandable, but neither names anything fundamentally new: peer review is already governed by incentives, such as reviewing others' submissions in return for having one's own work properly reviewed, and conferences already invest massive effort coordinating thousands of reviews and rebuttals with different rules and workflows. A credit system creates little out of thin air; it largely formalizes and aligns those incentives with the broader health of the ecosystem, and makes the existing effort fairer, more consistent, and more sustainable.\quad That said, we concede that the full system may be too heavy to implement at once, which is why we favor a gradual rollout of changes — a ``soft landing'' for existing community members, as we detail in Section~\ref{sec_rec_rollout}.

\subsection{``A credit system favors the privileged, and review duty exemptions would lose reviewers.''}
\label{sec_alt_point_exempt}

Some may argue that a credit system will disproportionately benefit researchers with more time, institutional support, or prior connections, allowing them to ``buy'' their way out of responsibilities (e.g., being exempt from review duties) while leaving others to ``pick up the slack.'' This is a legitimate concern, but in our design, points are earned through labor, not status. There is no ``premium tier of citizen,'' only accumulated contributions through hard work. While it is still true that researchers with strong support will likely have more opportunities to contribute (as they are not otherwise occupied by some chores), their ``surplus contributions'' are still a net gain to the community.

While exemptions from review duties might indeed cost us some reviewers, it is worth asking whether those willing to pay a high price to opt out are producing quality reviews (if kept by force), and whether they have the bandwidth to stay engaged with the authors. We tend to believe such answers lean toward the negative, and argue that a better alternative might be to just let them be exempted, then utilize the collected points to incentivize reviewers who do have the bandwidth and motivation in this particular cycle.

We would also strongly advocate \textbf{mobilizing researchers who are not main authors to participate more in the review pipeline}, as they likely have better bandwidth (since they are not under the pressure of author deadlines), and their reviews will not be as affected by feedback on their own submitted work. Under the current system, there is little incentive for researchers to do so, as most reviewers are recruited by mandatory reciprocity, which no longer applies without being an author. Our proposed credit system might provide them with a strong incentive to participate, as they can earn points to enrich their publication privileges; and specifically, have the option to spend such points to be exempt from reviewer duties when they are submitting lead-authored work, granting themselves wider bandwidth as authors when they are under rebuttal pressure.

\subsection{``What about early-career researchers?''}

Much like how games onboard novice players and companies onboard new employees, the point-hosting platform could grant a baseline amount of points to first-time contributors, perhaps along with a protection period, giving them enough time and capital for trial-and-error. Concretely, such a protection period (e.g., the first three to five months, or one's first few conference cycles) might allow a limited number of initial submissions to be fully refunded and non-extreme penalties to be softened, so that newcomers can learn from early missteps without those missteps derailing a nascent career. A central theme of the credit system is that a single mistaken penalty is not catastrophic; the same spirit should carry over to onboarding, where we can afford to be more forgiving.

We further note that a credit system can reward ``good acts'' well beyond reviewing itself. For instance, conference organizers might host onboarding workshops, or pair early-career researchers with more experienced ``research buddies'' who mentor them on submission compliance, review writing, and community norms. Rewarding such mentorship can be far more constructive than relying on point awards and penalties alone, yet would be hard to establish without a fine-grained credit system in place.

\subsection{``Voting-based penalties will be abused or weaponized.''}
\label{sec_alt_vote}

The voting-based lever introduced in Section~\ref{sec_proposal_defense} naturally raises the concern that it could be misused, weaponized in borderline cases, or influenced by interpersonal bias. This concern is legitimate. However, as argued in Section~\ref{sec_takes_reviewers}, the reviewing problems that matter most are the subtle ones that desk rejection is too blunt to police, which is exactly why a finer-grained, voting-based penalty is needed: if the authors report a reviewer and that reviewer's peers, along with the area chair, agree that a review is unacceptably low in quality or that the reviewer engaged in unprofessional conduct, the reviewer could face penalties ranging from a warning to graduated point deductions.

Such a system does introduce the possibility of false positives, but we argue this is largely acceptable. First, with guardrails such as close-unanimous (if not fully unanimous) agreement among the other reviewers on the paper, area chair confirmation, and an appeal mechanism, the practical false-positive rate can be kept low. Second, a point deduction is far less extreme than desk rejection or a submission ban, so even a wrong call is unlikely to cause severe or irrecoverable harm. Most importantly, the current system has the opposite problem: a nearly 0\% true-positive rate, since no matter how badly a reviewer behaves, there are essentially zero consequences short of automatically verifiable atrociousness. The status quo, in our view, is worse.

No penalty system will ever be perfect, but the absence of one leaves little room for improvement. We argue that a small risk of overcorrection is a worthwhile price for finally holding peer review to a higher standard. An even lower-risk alternative is to lean on the \emph{award} side of the same lever, granting credits to ACs and reviewers who provide detailed feedback on their peers' reviews, enabling a positive feedback loop where actors have a channel to learn and improve. We discuss such recommended practices in Section~\ref{sec_rec}.

\subsection{``Restricting point transfers is undemocratic.''}
\label{sec_alt_transfer}

It is worth first being explicit about why unrestricted transfer would break the system. Points are meant to certify that a specific person performed a specific service; once they can be gifted or sold, a researcher's privileges no longer reflect their own contributions, and the ``labor for perks'' principle the whole design rests on (Section~\ref{sec_proposal_defense}) loses its meaning; transferability would also neutralize much of our four-part defense, as a well-resourced actor could simply pool points from others to absorb the rising costs of \textit{dynamic pricing}. A transfer channel is, moreover, the natural on-ramp to a black market, leading to a money-to-points and pay-to-play outcome all reasonable scholars would want to avoid, and plausibly worse than what we have today.

Given this, one might still object that strictly regulating transfers is paternalistic, much like tightly regulating the transfer of money. We find the analogy imperfect: points are not private wealth but a record of personal community service, closer to a professional credential, a citation, or an authorship than to money. One cannot sell a degree or a reviewing record, and few would call that undemocratic. After all, non-transferability is what gives the credential its meaning, not a liberty taken away. The restriction on person-to-person transfer is also not a blanket ban on all sharing/pooling operations: team-based redemptions (e.g., coauthors jointly spending points) remain sensible, and each conference can set its own boundaries. What we resist is decoupling points from the labor that earned them.

\subsection{``Will points simply inflate as the system scales?''}
\label{sec_alt_scale}

A natural concern at scale is not the raw submission count but the points themselves: as the system runs across many cycles and venues, the total points in circulation could grow until they inflate and lose value as either a deterrent or a reward. This is a familiar problem for real-world point economies, and we can borrow their remedies. Credit-card and loyalty-point programs routinely curb inflation and hoarding through point expiration and time-limited, discounted redemption windows that nudge members to spend rather than stockpile, giving organizers meaningful influence over the effective point supply.\quad A second, related worry is that the system could be gamed at scale through cheap, bulk, or AI-driven point farming; this is contained by the four-part defense elaborated in Section~\ref{sec_proposal_defense}. None of this makes the system immune to abuse at scale, but it does hand organizers concrete moderation levers that current conference mechanisms simply lack.

\subsection{``A cross-conference reciprocity must exist first.''}

One clear and legitimate criticism of our credit system is that, for it to work to its full potential, multiple major conferences must adopt it. Granted, conferences like ICML, NeurIPS, and ICLR rarely collaborate explicitly, so this prerequisite is admittedly hard to meet. However, we argue that there are ML conferences well-positioned to adopt such practices: for example, the ARR series of conferences has long implemented cross-conference measures (e.g., submission bans from the next ARR cycle), as experimented with in \href{https://2025.emnlp.org/reviewer-policies/}{EMNLP 2025}, making them more openminded to adopting similar measures. Further, even if the credit system is per-conference, it can still function better than nothing; it is just that features requiring accumulated effort may be harder to activate and experiment with.

\subsection{``Who pays for the cost?''}

Our mechanism does not require conferences to offer more complimentary registrations than they already do. Major ML conferences already grant free registration to top reviewers.\footnote{NeurIPS 2025, for instance, offers this to an estimated \href{https://neurips.cc/Conferences/2025/ProgramCommittee\#top-reviewer}{1,900+ individuals}, and \href{https://x.com/icmlconf/status/2049919247065694669}{ICML 2026} similarly awarded free registration to its top 25\% of reviewers.} With conferences increasingly open to virtual attendance, the material cost per registration is further reduced. As discussed in Section~\ref{sec_rec}, point thresholds can be calibrated to match existing resource constraints.

The key difference is simply how these perks are allocated: instead of relying on AC discretion, we can rank reviewers by credits earned across multiple papers (or even venues), providing a more objective way to identify top contributors — and arguably giving people more freedom in how they spend their points.

\subsection{``Just enforce monetary cost per submission.''}

Some ML conferences have begun experimenting with monetary submission fees to curb submission volume.\footnote{\href{https://2026.ijcai.org/ijcai-ecai-2026-call-for-papers-main-track/}{IJCAI-ECAI 2026}, for instance, charges \$100 per paper from the second submission onward.} We argue that points are preferable to money for one practical reason: if the monetary threshold is too low, it becomes meaningless as a deterrent; if too high, it excludes researchers without strong institutional support — and blocking the accessibility of science is of course undesirable. Points, by contrast, can be earned by contributing to the community (e.g., by reviewing papers), making them a fairer currency that rewards effort rather than financial privilege. The restrictiveness of point transfers also makes them much more resistant to gaming and abuse, as we discussed in Section~\ref{sec_alt_transfer}.

\section{Recommended Practices / Call to Action}
\label{sec_rec}

As emphasized throughout this position paper, our goal is not to promote a single, prescriptive rulebook that every conference must follow, but to advocate for a flexible framework that can adapt to different conference idiosyncrasies. Under our credit system, conference panels and authors are akin to \textbf{store owners and customers: the panels decide what goods are offered and at what price, while the customers decide where and how they wish to spend their money.}\quad However, we recognize that without concrete discussion of how such a system might operate, adoption could invite resistance or, worse, chaos. This section therefore offers practical guidance on how the first few conferences adopting a credit-like system might proceed.

\subsection{Heavy on existing perks}

We believe that early adopters should anchor their credit-like system around perks ML conferences already offer (e.g., complimentary registration, emergency reviewer invitations), rather than immediately introducing entirely new perks (e.g., exemption from review duties, invitation of extra reviewers). Staying with existing perks provides two immediate benefits:
1) Because these perks are already part of established workflows, redistributing them according to the credit system (e.g., ranking reviewers by awarded points per conference cycle rather than relying on an AC's subjective judgment) keeps overall impact bounded. If the new distribution turns out problematic, its effects are still confined to the known scope of these already-tested perks; whereas brand-new perks introduce unknown risks.
2) We can directly compare credit-system conferences' key metrics against their historical data. Any improvement or decline is then more likely attributable to the credit system (or its specific implementation), rather than being confounded by the introduction of new perks. Reusing the same perks under credit-based allocation also yields ablated data on how the credit system behaves in practice — a baseline for testing new perks later.

\subsection{Gentle and gradual rollout of new perks, potentially with one-off tests}
\label{sec_rec_rollout}

When launching new perks, it is best to roll them out gently and gradually rather than all at once. This approach offers a clean testbed to monitor each perk's contribution and reduces the information load on all involved parties, who will need time to adjust.

Observant readers may notice that some of our proposed policies (e.g., free registration, the right to request additional reviewers, or refundable submission fees discussed in Section~\ref{sec_proposal}) require a relatively long-term accumulation of points before they become useful, stretching the evaluation horizon.

A simple way to expedite early evaluation is to introduce one-off tests. For example, in addition to point awards, conferences might grant top point-earners a one-off right to request an additional reviewer to help resolve borderline cases, with this right expiring at the end of the conference cycle. This allows organizers to directly observe whether such redeemable incentives meaningfully help, and to what extent these improvements propagate through the reviewer–AC pipeline. This kind of fast feedback might help conference organizers trim unhelpful policies quickly and enable faster iterations of rule sets.

\subsection{Determining point values for contributions and perks}

One reason we did not specify exact point values for different contributions (e.g., how many points an emergency review should yield) is that we currently lack the empirical data needed to set these responsibly. As a general guideline, we believe it is reasonable to treat the completion of one regular review duty as the base ``unit price'' of this ecosystem.

Rather than arguing directly about how many points each contribution ``should'' receive, we propose working backwards from the perks: estimate how many free registrations or similar rewards a conference can offer, determine what percentile of contributors this represents, and calibrate point values accordingly.

\subsection{Track key metrics and publicize such statistics}

Finally, for a credit system to have a lasting impact, conferences must make informed decisions about which rules to adopt and at what point-values. Such decisions require cross-conference consistency. If, for example, NeurIPS values its perks at 10x ICLR's level for no meaningful reason, the ecosystem loses the interoperability we envision. Thus, each conference should monitor key metrics and publish these statistics as part of their post-conference fact sheets.

For example: If a new rule is implemented, do we observe increased interaction among reviewers and ACs? Do ACs report that these additional exchanges help them make more confident decisions? These statistics and reports can form the foundation for iterating toward a better implementation of the credit system, and can serve as a strong signal, even an advertisement, encouraging more conferences to adopt a shared credit currency.

\section{Limitations}
\label{sec_limitations}

A proposal is only faithful if it also highlights its main weaknesses. In our case, \textbf{the lack of numerical experiment results} is one. However, we believe discussions about review mechanisms are most meaningful in the hypothetical space, since there is no way to rewind history and A/B-test different conference measures. LLM-powered simulation is a natural alternative, but at the scale of multiple conferences, we find it adds little value: too many moving factors compound, with too few publicly available metrics to ground them, making even crude simulations hand-wavy and easy to rig absent real baselines.\quad That said, recognizing the desire for some anchoring to real conferences, we share three case studies (from top ML venues where we served as reviewers) in Appendix~\ref{app_case_study} as a compromise between realism and faithfulness.

A second limitation is in terms of scope, as \textbf{our proposal is largely \emph{corrective}}: it polices and rationalizes an already-large pool of submissions, \textbf{rather than addressing the root cause behind that volume.} The explosion in ML submissions has turned review by a committee of experts into reliance on a committee of authors (often recruited by force), and we don't see this fundamental scaling challenge being addressed even with a credit system. This work thus leans toward the \emph{reactive} side of the problem — making review fairer and more sustainable given the volume we already face — rather than \emph{proactively} rethinking the publication system itself. Some of our measures do curb submission volume, but they remain closer to policing than to reducing the innate drive behind it; we regard the latter as a deeper question deserving dedicated study, and encourage future work in that direction.

\section*{Acknowledgements}

I thank the \textbf{reviewers and area chair} for their concrete and constructive feedback, and in particular for the many interesting threat models they raised, which shaped the four-part defense of the credit system. Many of the added alternative views were also inspired by their comments. Due to page limitations, I unfortunately could not include a diagram of the proposed pipeline in the main text as promised, but I will make sure one is available on my poster.

I would also like to thank \textbf{Moshe Y. Vardi} for his insightful comments on this paper, many of which I incorporated into the final version. Beyond individual fixes, Moshe pointed me to several relevant works I had overlooked, letting me attribute credit more properly — fitting, perhaps, for a paper that itself advocates crediting contributions. His observation that this proposal is more corrective than root-solving also prompted the scope discussion now in Section~\ref{sec_limitations}.

I am also grateful to \textbf{Buxin Su} — one of the core authors of the Isotonic Mechanism series of works \citep{su2025self_rank, su2026rejoinder} pioneered by Weijie Su \citep{su2021own_best_reviewer} — for an online discussion in which he shared feedback on my initial ideas. Many of this work's arguments were also challenged and polished by \textbf{Minghao Yan} and \textbf{Zhaozhuo Xu} when we drove back from NeurIPS 2025. I thank them for their company and sharp comments, and for letting me stay a passenger princess throughout the ride.

This work is written for the love of the game, out of a stubborn hope that ML review can be better: that one day a system — this one, something similar, or a different but better one — might provide an improved review experience for all actors in the pipeline. The nature of this work is not technical, and its development was not funded by any organization or grant.\quad That said, I would like to thank the \textbf{Department of Computer Science at Rice University} for providing a supportive research environment. On a similar note, much of the writing was done while I was visiting \textbf{MATS Research} and \textbf{Constellation (with  Anthropic Fellows Program)}, and I thank these organizations for their hospitality and for the chance to bounce ideas off my peers. In particular, I thank \textbf{Jonathan Michala} at MATS for being a wonderful research manager and for enduring much of my ranting and rambling about peer review. 

The views expressed in this paper are solely my own and do not reflect the views of any organization or any individual mentioned.


\bibliography{ref}
\bibliographystyle{icml2026/icml2026}

\newpage
\appendix
\onecolumn

\section{Related Works}
\label{app_related}

\paragraph{Proposal of conference mechanisms for better ML review quality} To the best of our knowledge, few published works have addressed \textbf{how to improve ML review quality by proposing new conference mechanisms.} The closest work to ours might be \citet{kim2025reviewer_rewards}, where the authors advocate establishing a feedback loop to reviewers and promoting reviewer rewards, two points that we also share. Specifically, \citet{kim2025reviewer_rewards} highlights that if we allow authors to rate reviewers, those ratings will almost always be heavily influenced by the specific strengths and weaknesses (and, by extension, the scores) listed by the reviewer \citep{goldberg2025peer_peer}. Reviewers who fairly rate papers negatively may be subjected to unfair retaliatory ratings from authors. To address this, \citet{kim2025reviewer_rewards} suggests a two-stage reveal: authors first read only the reviewer-written summary and strengths and provide a rating, after which the weaknesses are revealed. We argue that this system might work to some degree, but in reality, much of a review's quality is determined by whether the highlighted weaknesses are sound and well supported. Rating reviewers without seeing these details would likely produce noisy signals, and reviewers would be incentivized to write vague summaries that lack sharp substance; even worse, reviewers could inflate their strengths sections to manipulate the rating. \citet{kim2025reviewer_rewards} themselves acknowledge that their proposed mechanism cannot address low-quality negative reviews,\footnote{\url{https://openreview.net/forum?id=l8QemUZaIA&noteId=SXdGgGs6SV}} which are likely the main complaints of most authors.\quad As for reviewer rewards, \citet{kim2025reviewer_rewards} mostly argues for vanity perks like digital badges (e.g., ones similar to the ``Pull Shark'' badge on GitHub). We argue that such vanity-only perks would have much less influence than the review, submission, and cost-influencing perks we propose here.

Another piece of related work is the Isotonic Mechanism score pioneered by \citet{su2021own_best_reviewer} and its follow-up works like \citet{wu2023truth_serum, su2025self_rank}, which survey authors (with multiple submissions) and ask them to rank their submitted papers. A score is thereby calculated and compared with the mean of reviewers' raw scores. Should these two scores exhibit too drastic a gap, it may trigger AC intervention with an additional reviewer request, etc. We note that this work is, by and large, orthogonal to ours, as it is yet another safeguard one can implement under our proposed credit system. The two works overlap in the sense that some countermeasures do show resemblance (e.g., requesting an additional reviewer).

A broader but still relevant work is the survey by \citet{shah2022challenges_pr}, which systematically examines peer review challenges across several dimensions: mismatched reviewer expertise, dishonest behavior, miscalibration, etc. For each dimension, Shah discusses computational solutions, such as randomized assignment algorithms to mitigate dishonest bidding, or machine learning approaches to address commensuration bias. While comprehensive, the survey proposes separate solutions tailored to each problem rather than a unified mechanism. Our credit system, by contrast, offers a single flexible framework that can incorporate many of these solutions as special cases.

Two other pieces of related work are \citet{rogers2020peer_review_nlp} and \citet{zhang2022system_peer}. \citet{rogers2020peer_review_nlp} discusses why incentives clash under a peer-review context and outlines many potential proposals (e.g., better review–paper matching, more tracks, abolishing ``score-based'' feedback, track-specific review formats, etc.). Similarly, \citet{zhang2022system_peer} investigates various policies but focuses on modeling why resubmission is so prevalent despite many works eventually getting accepted at a top venue. Like \citet{shah2022challenges_pr}, these works propose custom solutions for each problem or edge case, rather than a unified recipe.

\paragraph{Credit-based review systems} A small but growing set of works, like ours, frames the remedy as a credit-, token-, or general reward-based currency for review. Most directly comparable in mechanism, though set outside ML, is the concurrent work of \citet{Francia2026peer_review_tokens}.\footnote{Which first appeared publicly after ours, despite an earlier initial submission.} Like ours, it holds that voluntary fixes fall short and builds the remedy on a credit-like currency for review; our differing settings, however, lead to substantially different designs along several axes:
\begin{itemize}[leftmargin=*, noitemsep, topsep=0pt]
\item \textbf{Venue Setting.} The proposed token system under \citet{Francia2026peer_review_tokens} is \emph{primarily} aimed at reducing review delay (shrinking editorial queues and submission-to-decision times) under constraints especially salient for scientific journals: reviewer scarcity, hard-to-place ``unattractive'' papers, and slow rolling turnaround. These are far less pressing at ML conferences, which draw on a comparatively ample reviewer pool and operate in fixed, batched cycles; our work's central problems are instead submission \emph{volume} and review \emph{quality and accountability}.\footnote{We want to be clear that \citet{Francia2026peer_review_tokens} also employ levers regarding review quality, though more at a conceptual level, with fewer operational proposals.} One of its core remedies, \emph{compulsory} reciprocity (``from volunteerism to duty''), is moreover already standard practice in ML and is, as argued in Section~\ref{sec_implicit_expectations}, regarded as problematic, plausibly because ML's far larger scale can turn compelled reviewing into many unwilling, ill-suited reviews that erode quality and are much harder to police.
\item \textbf{Penalty.} The scheme is almost entirely reward-\emph{reduction} rather than punishment: a late or weak review merely earns a fraction of a token, and a balance typically turns negative only for an \emph{undelivered} review. In ML, such a case is already egregious enough to warrant desk rejection, precisely the blunt instrument we argue is too coarse; it thus offers comparatively little for the subtle, long-tail misconduct that most needs policing, whereas we provide graduated, peer- and AC-adjudicated point \emph{deductions} spanning minor to severe.
\item \textbf{Reward.} The currency is largely single-purpose: one earns primarily by reviewing and spends primarily to submit, so it exists mainly to balance the review-to-submission loop. We instead treat reward as a general lever for shaping behavior across the pipeline, attaching points to many actions (e.g., emergency and ``outstanding'' reviews, initiating internal discussion, AC investigation, authors who return useful feedback to reviewers, and mentoring early-career researchers) and to many redemptions.
\item \textbf{Transferability.} The two designs largely agree: both let credits travel with the holder across venues, and both permit coauthors to pool credits toward a shared submission. They diverge on one point: \citet{Francia2026peer_review_tokens} additionally allow transfers between unrelated parties (e.g., a non-coauthor ``mentor'' may supply a rookie's tokens), which the balancing mechanism relies on but which, as the authors acknowledge, invites ``fake coauthorship.'' We forbid such person-to-person transfer, along with any money-to-points conversion, keeping points tied to the labor that earned them (Section~\ref{sec_alt_transfer}).
\end{itemize}
Both works engage a range of threat models and operational concerns; ours places somewhat more emphasis on defending against malicious gaming, through the four-part defense of Section~\ref{sec_proposal_defense}.

Outside the machine learning community, we also have \citet{gasparyan2015reward_peer} analyzing peer review incentives under a mainly medical-focused context. The main argument of this work is \textit{``none of these (financial or nonfinancial incentives) is proven effective \textbf{on its own}''}; however, the authors envision \textit{``a strategy of combined rewards and credits for the reviewers' creative contributions seems a workable solution.''}\quad Our work makes essentially the same argument, though framed under a credit system and tailored to the ML setting. Our system supports many of the typical financial and non-financial incentives mentioned in \citet{gasparyan2015reward_peer}. For instance, several exemplary policies we discuss in Section~\ref{sec_proposal_points} range from incentives for reviewers to write higher-quality and more timely reviews, to deterrents for authors submitting unready work, to non-financial privileges such as the right to be exempt from review assignments, and even financial compensation like free registration. We do not advocate for any particular type of incentive, but rather a collection of them, unified under a credit system so that their impact can last beyond a single conference.

Another point specific to \citet{gasparyan2015reward_peer} is that many reviewers dismiss incentives such as paper purchase discounts or free publication access, since their institutions already pay for publisher subscriptions, making such incentives mostly relevant to members with non-academic backgrounds. In our proposal, however, conference hosts can offer services that no institution would purchase in bulk (e.g., registration), or even privileges that money cannot buy (e.g., the right to request additional review resources). While we do not claim these incentives are inherently better, we do believe our framework offers a flexible way to combine different incentives to suit diverse needs, effectively pushing forward a system that \citet{gasparyan2015reward_peer} envisions.

Real-world platforms such as \texttt{ReviewerCredits} \citep{fund2024reviewer_credits} and \texttt{Publons} \citep{publons2018global_state} have also explored giving reviewers some form of recognition or credit for their review activity. Such efforts broadly share our premise that review labor merits durable, portable credit, though they generally operate as voluntary, third-party services rather than mechanisms tied to a venue's own decisions. This is part of why we instead tie credit to concrete stakes within the conference pipeline.\quad Related ideas also surface outside academia, in the opensource community. For instance, \texttt{vouch}\footnote{\url{https://github.com/mitchellh/vouch}} implements a ``web of trust'' in which members vouch for (or denounce) one another to manage participation. While conceptually adjacent to our credit system, the resemblance is loose: \texttt{vouch} is largely focused on access control, gating who is admitted in the first place, whereas our proposal is aimed at shaping behavior after admission (for obvious fair-science reasons), with greater flexibility in how earned points can be spent. We thus see it as related in spirit but distinct in goal.

\paragraph{Conference Review Statistics} There are a few works like \citet{yang2025paper_copilot, cortes2021neurips_consistency_2014, beygelzimer2023neurips_consistency_2021, goldberg2025peer_peer} that collect real statistics and conduct controlled experiments from past conferences. While such works typically do not propose mechanism-based solutions, their numerical presentations help illustrate the scale and disorder of current conferences; nonetheless, such controlled experiments shall offer us insight into the practical dynamics of a particular mechanism design.

\paragraph{Broader Relevant Art} Beyond the effectiveness of any single incentive system, much of a conference's experience also depends on many other aspects, such as the reduction of collusion rings \citep{jecmen2025collusion_ring}, finding better reviewer–paper mappings \citep{mimno2007reviewer_paper_mapping}, determining the level of openness \citep{rao2025openness_peer_review}, or the use of LLMs for reviewers \citep{liu2023reviewergpt}. We refer readers to these works to build a deeper understanding of these important topics, and recommend \citet{kim2025reviewer_rewards} for an overview of such efforts.

\section{Three Case Studies Where We Served as Reviewers}

\label{app_case_study}

While our position paper, unfortunately, lacks real conference data to support why our proposed framework would be helpful, we believe there is even less reason to run an LLM-roleplaying simulation. We understand the perspective that having some anchors to real conferences is preferred. Here, we share three case studies, all from similar top ML conferences, where we served as reviewers. We share these not to highlight our own conduct, but to illustrate how small reviewer initiatives can help.

\subsection{Case 1: Reviewers criticizing matters outside the paper's scope.}

In this case, we observed that another reviewer criticized the submitted work for reasons clearly outside its intended scope. We therefore raised our concern to the AC:

\begin{tcolorbox}[colback=lightgray!10, colframe=black, title={Internal comment to PC/SAC/AC}]

This message is set to be only visible to PC, SAC, AC, and the authors.

I want to disclose that I find reviewer \texttt{A}'s evaluation of this paper quite unreasonable. This reviewer writes:
\begin{itemize}
    \item Certain methods, such as \texttt{method type}, demonstrate limited effectiveness in \texttt{setting}, which may restrict their practical deployment.
    \item The paper points out that many of the \texttt{an important task} methods tested are essentially extensions of existing models adapted for \texttt{another important task}, such as \texttt{a famous method}.
\end{itemize}

It doesn't make much sense to cite the low performance of certain featured methods as weaknesses of a dataset-proposing/benchmark paper. It is not the authors' problem if an established method underperforms. Instead, the point of benchmarking is precisely to show when a method would fail. Many benchmark works have done this --- \texttt{some examples} --- and it is incomprehensible why this is considered a weakness.

Another criticism from reviewer \texttt{A} is:
\begin{itemize}
    \item The tasks within the benchmark may not capture all possible real-world application scenarios, possibly overlooking specific needs within certain domains.
\end{itemize}

This, in my opinion, is a boilerplate concern that can be said for literally \emph{any} dataset. While I do agree that the proposed dataset does not capture some important \texttt{task} scenarios --- \texttt{some examples} --- criticizing it for ``not capturing all possible real-world applications'' crosses the line and feels borderline hostile. This is akin to criticizing a method paper for not evaluating on every possible dataset.

I recommend the AC to either disregard \texttt{A}'s review or consider encouraging the reviewer to revisit the evaluation.
\end{tcolorbox}

This paper was ultimately accepted. This anecdote shows that without internal reviewer analysis, simple rule-based policies such as ``inactive $\rightarrow$ desk rejection'' fail to capture cases of severely low-quality reviews. Finer-grained measures must be in place to ensure that positive impact scales broadly, rather than being limited to a few desk-rejected papers.

\subsection{Case 2: Reviewers asking for particular experiments after the rebuttal deadline.}

In a top ML conference where the exchange between authors and reviewers is limited to a certain time window, we had a split decision situation where the non/late-responding reviewers were not supportive of the submission. As the AC was calling for consensus, we stepped in and asked:

\begin{tcolorbox}[colback=lightgray!10, colframe=black, title={Internal reviewer discussion}]
I skimmed over the two negative reviews of this work and found merits in many of the reviewer-raised points. However, I also find the authors' rebuttal to be proper in many regards — especially when the raised concern demands a clarification-like answer.

It looks like the two negative reviewers have yet to address the authors' rebuttal in a meaningful way (only acks are issued, cmiiw). So, to reach a consensus, I believe it would be helpful if the two reviewers could elaborate a bit on their leftover concerns. I am happy to set aside some time to discuss such leftover issues from my perspective.

\end{tcolorbox}

Essentially, the two negative reviewers believed that certain experiments were missing — one of which can be seen as a combination of two existing methods, and another as a specific investigatory study of the author-proposed method. While we find such suggestions to have merit, we believe they were not raised appropriately from a procedural standpoint:

\begin{tcolorbox}[colback=lightgray!10, colframe=black, title={Internal reviewer discussion}]

I appreciate \texttt{A}'s detailed response and updated review. I believe \textbf{\texttt{A} (as well as \texttt{B})'s main concerns regarding \texttt{ProposedMethod} vs.\ \texttt{PriorWork1} + \texttt{PriorWork2} are legitimate and sound.} That being said, I am always the kind of reviewer who is ``more in the authors' shoes''---for lack of better words---and I would like to present two alternative arguments regarding this concern.

First, I believe experiment-comparison requests that touch on \emph{combinations of existing works} should be cautiously brought up. Many methods can be combined, but their combinations typically require a number of discretionary design decisions, and it is often unlikely for authors to feature the exact combination a reviewer has in mind. In this case, \texttt{ProposedMethod} proposes a paradigm of \texttt{[redacted]}, where the scope of eligible combinations is wide. Thus, in my opinion, \textbf{if reviewers are specifically interested in the comparison \texttt{ProposedMethod} vs.\ \texttt{PriorWork1} + \texttt{PriorWork2}, such a request should be made \emph{explicitly} before the rebuttal deadline, rather than mentioned in hindsight when the authors have no channel to address it.}

\medskip

From the look of it, the \texttt{ProposedMethod} authors submitted their initial rebuttal on \texttt{an early date}, which is \texttt{[redacted]} days after the review post. However, only \texttt{A} engaged substantively on \texttt{a late date}. I must note that this year's \texttt{BigConferenceName} requires only two rounds of exchange, yet only 2/4 reviewers provided those to the authors, with all engaged reviewers leaning positive. \textbf{For such reasons, while I am also interested in this comparative result and agree with \texttt{A}'s analysis, I do not believe we can use it against the authors (at least not as a singular veto reason), as the request was not properly raised from a procedural standpoint.} Imagine we were submitting a paper where reviewers were largely non-responsive, and the paper was then rejected for missing an experiment that was never explicitly asked for---it would be hard not to feel that is unfair. While I understand we all have different priorities and may have limited bandwidth for various reasons, we need to properly compensate authors when such situations occur.

\medskip

(I know it is uncommon for a reviewer to argue on behalf of authors, but I always do so when it is warranted. The AC is welcome to confirm that this advocacy is from me for good reasons and not the result of any collusion.)

\end{tcolorbox}

Later, we and the two reviewers exchanged more than five comments in total, which contributed to the AC's eventual decision. This anecdote shows that many reviewers are willing to engage in internal discussions, and such exchanges are profoundly helpful in deciding borderline papers — they simply never had the ``push'' to initiate such discussion voluntarily. We argue that our credit system could help elicit more of these productive discussions.

Another takeaway from this exchange is the importance of having enforceable safeguards (e.g., reply deadlines), as otherwise authors may have no channel to meaningfully rebut at all. Attaching rewards and penalties to such actions is also crucial to ensure procedural fairness. While we respect and appreciate \texttt{A} and \texttt{B} for engaging in our discussion, they were fine with leaving authors unanswered or replying late only because there is virtually no penalty to them (as their ``misconduct'' would be too minute for desk rejection, yet the conference organizer has no other finer-grained tools) — something our credit system would help address.

\subsection{Case 3: Authors presenting unsupportive results while claiming otherwise.}

In this case, we found that the authors were presenting unsupportive results to one reviewer while verbally claiming the opposite — so we stepped in:

\begin{tcolorbox}[colback=lightgray!10, colframe=black, title={Internal reviewer discussion}]
As another reviewer, I find the reading on \texttt{ProblemX} tricky. The goal of \texttt{Task} is to have the \texttt{[redacted]}.

In \texttt{ProblemX}, when the \texttt{component} is trained only on \texttt{dataset1}, the \texttt{dataset2} accuracy drops quite significantly — a disadvantaged result. Once the \texttt{component} is trained on both \texttt{dataset1} and \texttt{dataset2}:
\begin{itemize}
    \item The \texttt{dataset1} performance improves \texttt{a small number} over the \texttt{a baseline}, but this also comes at the cost of generating many more tokens than the \texttt{a baseline}.
    \item \texttt{[redacted as it is too technical]}
    \item In the end, the system shows only a \texttt{small number} accuracy gain on the task it was specifically trained on, while \texttt{doing something at a higher cost}. I think the added experiment makes the work more negative (though I appreciate the transparency), as it feels like the proposed \texttt{component} is very task-specific and does not generalize well.
\end{itemize}

(This message is set to be not visible to authors so that we can have a discussion with supposedly no bias, but I am happy to adjust the visibility should you want a more public discussion.)
\end{tcolorbox}

The reviewer exchanged views with us and we reached an agreement. The paper was ultimately rejected, with the AC citing concerns raised during the discussion. Specifically, this other reviewer noted:

\begin{tcolorbox}[colback=lightgray!10, colframe=black, title={Internal reviewer discussion}]
(BTW, you're among the very few reviewers who respond to other review comments. I truly admire this level of dedication.)
\end{tcolorbox}

This suggests that internal reviewer discussion is indeed rare, even though we lack direct statistical evidence.

\medskip

We hope these three anecdotal case studies illustrate how even a small component of our proposed framework (encouraging more internal reviewer discussion) can meaningfully facilitate paper evaluation. While we acknowledge our bias, we believe that in all three cases, our initiative in starting these discussions clearly shaped the outcome. This suggests that most reviewers \emph{want} to contribute and most ACs will take such discussions seriously; they simply need a small push to take the initiative — a push that our credit system could very well provide.


\end{document}